\documentclass[runningheads]{llncs}
\usepackage{booktabs}
\usepackage{graphicx}
\usepackage{orcidlink}
\usepackage{caption}
\usepackage{subcaption}
\usepackage{algorithm}
\usepackage{algcompatible}

\algdef{SE}[SUBALG]{Indent}{EndIndent}{}{\algorithmicend\ }%
\algtext*{Indent}
\algtext*{EndIndent}

\begin{document}
\title{TinyMetaFed: Efficient Federated Meta-Learning for TinyML}
%
%
\author{Haoyu Ren\inst{1, 3}\orcidlink{0000-0002-0241-6507} \and
Xue Li\inst{2}\orcidlink{0000-0002-4515-6792} \and
Darko Anicic\inst{1}\orcidlink{0000-0002-0583-4376} \and
Thomas A. Runkler\inst{1, 3}\orcidlink{0000-0002-5465-198X}}
\titlerunning{TinyMetaFed: Efficient Federated Meta-Learning for TinyML}
\authorrunning{H. Ren et al.}
%
\institute{Siemens AG, Otto-Hahn-Ring 6, 81739 Munich, Germany\\
\email{\{haoyu.ren, darko.anicic, thomas.runkler\}@siemens.com}\\
\and
The University of Queensland, Brisbane, Queensland 4072, Australia 
\email{xueli@itee.uq.edu.au}\\
\and
Technical University of Munich, Arcisstr. 21, 80333 Munich, Germany
}
\maketitle              
\begin{abstract}
The field of Tiny Machine Learning (TinyML) has made substantial advancements in democratizing machine learning on low-footprint devices, such as microcontrollers. The prevalence of these miniature devices raises the question of whether aggregating their knowledge can benefit TinyML applications. Federated meta-learning is a promising answer to this question, as it addresses the scarcity of labeled data and heterogeneous data distribution across devices in the real world. However, deploying TinyML hardware faces unique resource constraints, making existing methods impractical due to energy, privacy, and communication limitations. We introduce TinyMetaFed, a model-agnostic meta-learning framework suitable for TinyML. TinyMetaFed facilitates collaborative training of a neural network initialization that can be quickly fine-tuned on new devices. It offers communication savings and privacy protection through partial local reconstruction and Top-P\% selective communication, computational efficiency via online learning, and robustness to client heterogeneity through few-shot learning. The evaluations on three TinyML use cases demonstrate that TinyMetaFed can significantly reduce energy consumption and communication overhead, accelerate convergence, and stabilize the training process.

\end{abstract}
\section{Introduction}

Over the past decade, the advancement of ML applications has been propelled by the emergence of big data and enhanced computational capabilities. This has led to a surge in large-scale AI models, such as "ChatGPT," which demands extensive resources and significant power consumption. The growing awareness in the ML community emphasizes the escalating resource requirement and environmental unsustainability associated with big AI models.

Tiny Machine Learning (TinyML) has emerged as a powerful paradigm that bridges the gap between ML and embedded systems. It brings real-time AI capabilities closer to the edge, shifting data processing from data centers to Internet of Things (IoT) devices. TinyML offers sustainability, data privacy, and efficiency advantages by minimizing the need for cloud data transmission. The current estimate suggests that over 250 billion IoT devices are in active use today, with continual rising demand, particularly in the industrial sector~\footnote{\url{https://venturebeat.com/ai/why-tinyml-is-a-giant-opportunity/}}. Considering the vast deployment of embedded devices, the question arises: can TinyML applications benefit from sharing and integrating insights gained from these devices?

Federated learning (FL) offers a distributed ML schema where clients collaborate to train a global model by merging their local updates on a central server without sensitive data leaving devices. However, studies have shown that training a common global model is not always optimal due to non-Independent and Identically Distributed data among devices~\cite{Zhao2018}. The complex and ever-changing deployment environment of TinyML, coupled with the distributed nature of IoT devices, leads to heterogeneous data distribution. A global model trained by FL may exhibit arbitrary performance degradation when applied to a new device. Additionally, each device may have different objectives for its ML task, such as different output classes of interest. The resource constraints and limited availability of labeled data on tiny devices further exacerbate the challenges.

\begin{figure*}[!t]
\centering
\includegraphics[width=0.95\textwidth]{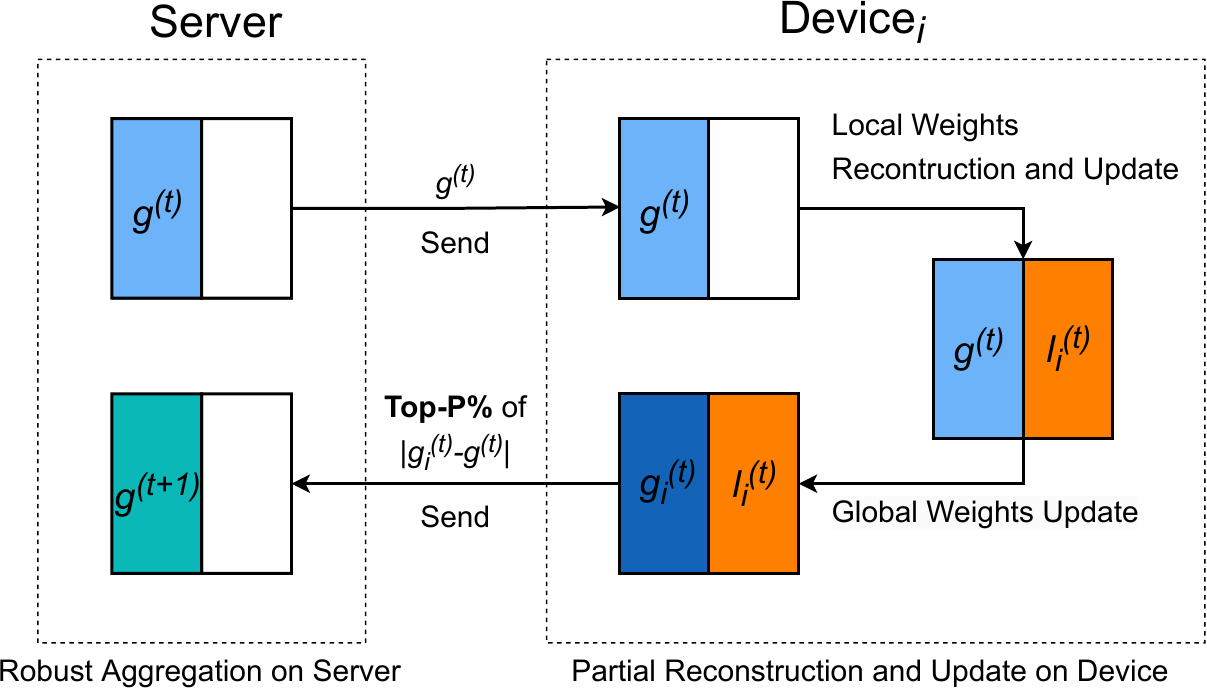}
\caption{Illustration of the TinyMetaFed workflow. Model weights are divided into global weights $g$ and local weights $l$. In each round $t$, a device $i$ use the received global weights $g^{(t)}$ to reconstruct its local weights $l_i^{(t)}$ and updates the global weights subsequently. In this work, a few gradient descent steps perform local weights reconstruction and global weights update. Then, the P\% updated global weights $g_i^{(t)}$ with biggest absolute changes $|g_i^{(t)} -  g^{(t)}|$ are sent back to the server. Finally, the server aggregates the Top-P\% updates to the central global weights using learning rate schedule with cosine annealing.}
\label{fig_1}
\end{figure*}

To address these challenges, we present TinyMetaFed in this work, depicted in Fig.~\ref{fig_1}. We consider a group of devices, each assigned an individual ML task from a distribution of tasks, e.g., one device classifies “dog vs. cat” while another classifies “apple vs. pear.” TinyMetaFed learns a Neural Network (NN) model initialization that can be quickly fine-tuned on a new device for its unseen task drawn from the same distribution, e.g., to classify “car vs. airplane.” The framework works by iteratively sampling a device, training on its specific task, and moving the initialization toward the trained model on that device. We explore partial local reconstruction to enhance communication efficiency and client privacy where a model is partitioned into global and local parameters. During each round, only the global parameters are communicated between a device and the server, ensuring that local parameters never leave the client. We further reduce communication costs by introducing Top-P\% selective communication, where only the P\% global parameters with the biggest changes are transmitted to the server. To address resource constraints, we propose online learning in TinyMetaFed, enabling on-device data processing in a streaming fashion. Online learning allows local models to process incoming data as it arrives without storing historical data, aligning with real-world production scenarios. Additionally, we improve model generalization performance by applying learning rate scheduling with cosine annealing.

\begin{figure}[!t]
\centering
\includegraphics[width=0.75\columnwidth]{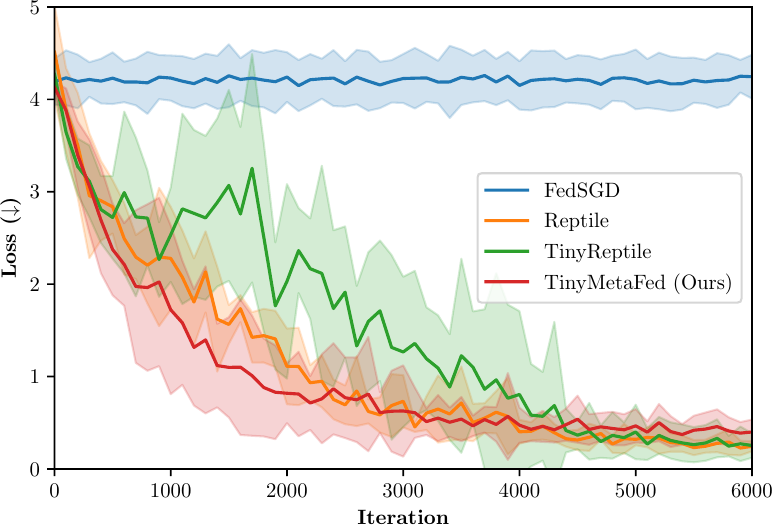}
\caption{Demonstration of FedSGD, Reptile, TinyReptile, and TinyMetaFed on the Sine-wave regression example. The model consists of five fully connected layers: 1 → 16 → 16 → 16 → 1. TinyMetaFed attains similar convergence performance as Reptile. However, TinyMetaFed requires significantly lower communication and computational costs per iteration than all other methods, as explained in Section~\ref{section:experiments}. While TinyReptile occasionally becomes unstable, TinyMetaFed consistently achieves faster convergence with comparable performance. Besides, traditional FL algorithms like FedSGD struggle with meta-learning settings.}
\label{fig_2}
\end{figure}

We evaluate TinyMetaFed on three meta-learning datasets: “Sine-wave example” for regression problem, “Omniglot” for image classification, and “Keywords Spotting” for audio classification. Our results demonstrate the superior convergence speed, computational and communication efficiency, and robustness of TinyMetaFed compared to the state-of-the-art methods while revealing the limitations of traditional FL architectures when dealing with heterogeneous local data, as shown in Fig.~\ref{fig_2}. 

The rest of the paper is structured as follows: Section~\ref{section:related_work} covers related work on FL, meta-learning, and online learning. Section~\ref{section:approach} presents the methodology of TinyMetaFed. Section~\ref{section:experiments} introduces the benchmarking datasets, describes the experimental settings, and analyzes the results. Finally, Section~\ref{section:conclusion} concludes the paper and explores future research directions.

\section{Related Work}
\label{section:related_work}

\subsubsection{Federated Learning}
FL focuses on training a NN model collaboratively across distributed clients whilst preserving local data privacy. Initial research on FL emerged in 2016, introducing baseline algorithms such as "FedSGD~\cite{McMahan2016}." Since then, numerous advancements have been made, including differential privacy~\cite{Jiang2023}, robustness~\cite{Aramoon2021}, optimization algorithms~\cite{Singhal2021}, and communication efficiency~\cite{Ji2021}. Although some studies have explored FL on edge devices like Raspberry Pi~\cite{Gao2020}, limited attention has been given to applying FL to the TinyML domain. State-of-the-art is limited to a few works~\cite{Kopparapu2022,LlisterriGimenez2022}, which have conducted experiments in simulation or controlled environments on a small scale, failing to address real-world challenges in TinyML~\cite{Imteaj2022}, such as client heterogeneity.

\subsubsection{Meta-learning}

Meta-learning is an appealing approach for rapid adaptation with minimal data. Unlike transfer learning, which pre-trains a global model and fine-tunes it on small datasets without guaranteed generalization~\cite{Parnami2022}, meta-learning aims to train a common model explicitly for easy fine-tuning. The initial advancement in meta-learning, known as "model-agnostic meta-learning" (MAML)~\cite{Finn2017}, introduced gradient-based optimization. However, MAML's requirement for higher-order derivatives makes it computationally intensive. Subsequent research focused on improving performance, such as MAML++~\cite{Antoniou2019}, and reducing the computational costs~\cite{Nichol2018}. 
Relevant work is emerging in the TinyML domain as well~\cite{Ren2023}. Nevertheless, most approaches require specific setups applicable only to certain models~\cite{Zhou2019} or remain computationally intensive~\cite{Rusu2018}, making them unsuitable for tiny devices with severe resource constraints.

\subsubsection{On-device Learning}
Most ultra-low-power TinyML models have traditionally been trained offline, while edge devices solely perform inference. Although this fashion has demonstrated success across the research community~\cite{Dhar2021} and industry~\cite{Montiel2020}, it is increasingly aware that offline-trained models may not be effective in real-world scenarios if data distributions evolve~\cite{Avi2022}. We believe that on-device learning is crucial for TinyML applications, allowing algorithms to adapt to heterogeneous deployment environments. Some research efforts have been dedicated to this field. For example, 
J. Lin et al.~\cite{Lin2021} have applied quantization-aware scaling and sparse update to train a convolutional NN under 256KB of memory. Compared to batch training, online learning has received less attention due to the assumption that data are always available as a batch. By integrating online learning into existing algorithms, we process sensor data one by one and update the model in a streaming fashion, saving computational resources and accelerating the training process. TinyOL~\cite{Ren2021} incorporates online learning to update the last few layers of a NN against incoming sensor data, keeping the model up to date without saving historical data. Our work combines meta-learning with online learning to enable on-device meta-learning across tiny federated devices.


\section{Method}
\label{section:approach}

\subsection{Federated Meta-learning}

In this study, we focus on federated meta-learning. Given limited training data, meta-learning enables a NN to adapt to new tasks or environments quickly. Federated learning employs distributed clients to collectively train a NN by communicating with a central server. Our approach combines meta-learning and federated learning to achieve meta-learning in a federated setting, leveraging resource constrained IoT devices.

Our setup consists of a server and a set of devices. Each device is assigned an ML task $t$ drawn from a distribution of tasks $T$. While all the tasks share a common pattern with the same number of output classes, such as two-classes image classification, their classification objectives differ. For instance, one task is to classify “dog vs. cat,” while another is to classify “apple vs. banana.” All devices are deployed with a NN of the same structure for their tasks. For evaluation, we divide these tasks into training tasks $T_{training}$ and testing tasks $T_{testing}$. Our algorithm utilizes $T_{training}$ to find an optimal model initialization $\phi$ that yields good generalization performance on $T_{testing}$, which have not been encountered during training. 


\subsection{TinyMetaFed}
This section presents TinyMetaFed, a framework that enables federated meta-learning in TinyML. We also discuss several techniques implemented within TinyMetaFed. Fig.~\ref{fig_1} provides an overview of TinyMetaFed. 

\subsubsection{Partial Local Reconstruction}
In federated learning, transmitting all model weights between devices and a central server can be communication intensive. We propose partial local reconstruction to address the challenge, where models are partitioned into local and global weights. The partition between local and global parameters depends on the use case requirements, privacy needs, and communication limitations. Clients only communicate their global weights with the server while preserving the local weights across iterations by recovering them whenever needed. In each round, a client receives global weights from the server and recovers its local weights through a few gradient decent steps. At the end of a round, the global weights are sent back to the server for aggregation, while local weights can be discarded or retained for local inference.

\subsubsection{Top-P\% Selective Communication}
Instead of sending all the global weights from clients to the server or randomly selecting some of them for transmission, we propose Top-P\% selective communication. Here, we assess the importance of each model weight. The approach selects the P\% global weights with the largest absolute changes and their indices for transmission in each round.

\subsubsection{Online Learning}
TinyMetaFed incorporates online learning in local weights reconstruction and global weights update, sequentially processing incoming data, such as sensor data, without storing them. This technique helps to minimize memory usage and keep the models up to date, which differs from traditional batch learning, where models are trained on entire stored datasets in batches.

\subsubsection{Learning Rate Scheduling
with Cosine Annealing}
Traditionally, a fixed learning rate is applied for training optimizers. Annealing the learning rate has proved crucial for achieving state-of-the-art results~\cite{Loshchilov2017}. Thus, we employ learning rate scheduling with cosine annealing to enhance generalization performance without extensive hyperparameter tuning.

\begin{algorithm}[htbp]
  \caption{TinyMetaFed}

  \begin{algorithmic}[1]
    \REQUIRE set of clients with tasks drawn from $T$, set of tasks $T$ with streaming data $D = \langle S, Q \rangle$, server learning rate scheduling function $f$.
    
    \ENSURE global weights initialization $g$

    \STATE \textbf{ServerUpdate:}
    \Indent
    \STATE Randomly initialize global weights $g$
    \FOR{each round $t$} 
    
    \STATE Sample one available client $i$ with $D_i = \langle S_i, Q_i \rangle$
    \STATE $ g_i^{(t)} \leftarrow  $ \textbf{ClientUpdate}$(i, g^{(t)})$
     \STATE $g^{(t+1)} \leftarrow  g^{(t)} + f(t) (g_i^{(t)} - g^{(t)})$
    \ENDFOR
    \STATE Return $g$
    \EndIndent
    \STATE \textbf{ClientUpdate:}
    \Indent
    \STATE $D_i = \langle S_{i}, Q_{i} \rangle, g^{(i)}$
    \STATE Freeze $g^{(i)}$
    \STATE $l_i^{(t)} \leftarrow  $ \textbf{LocalWeightsReconstruction}$(S_{i}, g^{(t)})$
    \STATE Unfreeze $g^{(i)}$ and freeze $l_t^{(i)}$
    \STATE $g_i^{(t)} \leftarrow  $ \textbf{GlobalWeightsUpdate}$(Q_{i}, l_i^{(t)}, g^{(t)})$
    \STATE Send \textbf{Top-P\%} of $g_i^{(t)}$ with the largest changes $|g_i^{(t)}-g^{(t)}|$ back to the server
    \EndIndent
    \STATE \textbf{LocalWeightsReconstruction/GlobalWeightsUpdate:}
    \Indent
    \STATE Perform $k$ steps of SGD on the streaming data in an online learning way
    \EndIndent
  \end{algorithmic}

  \label{a1}
\end{algorithm}

\subsubsection
TinyMetaFed, depicted in Algorithm~\ref{a1}, is inspired by the well-known meta-learning algorithm Reptile~\cite{Nichol2018}. The training process in TinyMetaFed operates as follows: in each round $t$, the server sends the current global weights $g^{(t)}$ to a client $i$. The client possesses a local dataset $D$, such as sensor data, specifically for its ML task. The dataset is split into two parts: a support set $S$ and a query set $Q$, where $D = \langle S, Q \rangle$. The client uses $g^{(t)}$ along with $S$ to reconstruct the local weights $l_i^{(t)}$. Afterward, the client freezes $l_i^{(t)}$ and produces updated global weights $g_i^{(t)}$ using $g^{(t)}$, $l_i^{(t)}$ and $Q$. The server then receives the Top-P\% updated global weights with the largest absolute changes $|g_i^{(t)}-g^{(t)}|$ from the client. Finally, the server applies the learning rate scheduling strategy to aggregate these updates into the global weights $g^{(t+1)}$. 

In this framework, local weights reconstruction and global weights update are achieved through $k$ local gradient descent steps in an online learning manner. The training processes are independent of the federated process, enabling clients to reconstruct weights once, store them for inference, and optionally refresh them with new local data. Our framework allows devices to communicate sequentially with the central server without relying on consistent and concurrent connections. Meta-learning strives to find a fast learner, so local datasets $D = \langle S, Q \rangle$ typically contain a limited number of samples, and the training step $k$ is generally defined to a small value. Intuitively, TinyMetaFed aims to bring a model initialization closer to an optimal point nearest to all tasks/devices, facilitating rapid fine-tuning in new environments. Essentially, TinyMetaFed optimizes for generalization. 


\section{Experiments and Evaluation}
\label{section:experiments}

In this section, we assess the performance of TinyMetaFed on regression, image classification, and audio classification tasks. We analyze various performance metrics, such as energy consumption, communication costs, and memory requirement. The NNs from the MLPerf Tiny benchmark~\cite{Banbury2021} are used to ensure consistent comparisons. Table~\ref{tab_1} provides an overview of the models for the three tasks. 
Each experiment is repeated many times, and the results are presented as the mean value with the standard deviation.

\begin{table}[tbp]
\centering
\caption{Overview of the models.}
\label{tab_1}
\begin{tabular}{@{}lllll@{}}
\toprule
                        & Task                 & Model Type      & Size    & Parameters \\ \midrule
Sine-wave Example       & Regression           & Fully Connected & 4.1 KB & 593        \\
Omniglot 5-classes          & Image Classification & Convolutional   & 313 KB  & 80389      \\
Keywords Spotting 3-classes & Audio Classification & Convolutional   & 71 KB & 17251      \\ \bottomrule
\end{tabular}
\end{table}

\subsection{Datasets}

\subsubsection{Sine-wave}
The Sine-wave regression task fits a randomly parameterized sine function ${f(x) = a \sin(b\, x + c)}$ on each device. The objective is to collectively learn a NN initialization that can be rapidly adapted to new sine functions using a handful of sampled pairs $(x, y)$.

\subsubsection{Omniglot}
The Omniglot image dataset consists of $C=1623$ characters from 50 alphabets, with 20 samples per character. In the meta-learning setting, each device is assigned a classification task of $M$ randomly selected characters from $C$. The goal is to learn good initial weights that can be quickly generalized to new devices with unseen classification tasks based on limited data examples.

\subsubsection{Keywords Spotting}
The Keywords Spotting dataset contains $C=35$ distinct keywords, such as "up" and "down," with over 1,000 audio samples per word. Each device is assigned a classification task involving $M$ randomly chosen keywords. The goal is similar to Omniglot, aiming to collaboratively learn a NN initialization that can be fast adapted to a new device with a different keywords classification task of $M$ classes given limited data.

\subsection{Baselines and Setup}

We compare TinyMetaFed with two state-of-the-art methods, Reptile~\cite{Nichol2018} (serial) and TinyReptile~\cite{Ren2023}. Reptile, MAML~\cite{Finn2017}, and MAML++\cite{Antoniou2019} are widely recognized as leading approaches in meta-learning. Compared to MAML and MAML++, Reptile stands out for its simplicity and efficiency. It achieves the goal by repeatedly optimizing the model initialization for different tasks and gradually updating the parameters toward the weights learned for new tasks. Our previous work, TinyReptile, applies the online learning concept to Reptile. This allows TinyReptile to process local data in a streaming fashion without the need to store past data, which saves a significant amount of resources and enables meta-learning on constrained devices. TinyMetaFed builds upon TinyReptile with various improvement strategies introduced in Section~\ref{section:approach}. We exclude other FL and meta-learning algorithms in the experiments since most of them are ineffective or unsuitable in the context of TinyML. 
We experiment with different hyperparameters that perform well for the given datasets: the Top-P percentage (10\%-80\%), SGD learning rate $\beta$ (0.001--0.02), support set size $S$ (1--16), query set size $S$ (1--16), and partitioning of global and local weights. Although, we do not fine-tune them for optimal results. 
The evaluation is conducted on Arduino Nano BLE 33 microcontroller (MCU) ~\footnote{\url{https://docs.arduino.cc/hardware/nano-33-ble-sense}} and Raspberry Pi 4 Model B~\footnote{\url{https://www.raspberrypi.com/products/raspberry-pi-4-model-b/}}.

\subsection{Results}

\begin{figure}[tbp]
     \centering
     \begin{subfigure}[b]{0.48\textwidth}
         \centering
         \includegraphics[width=\textwidth]{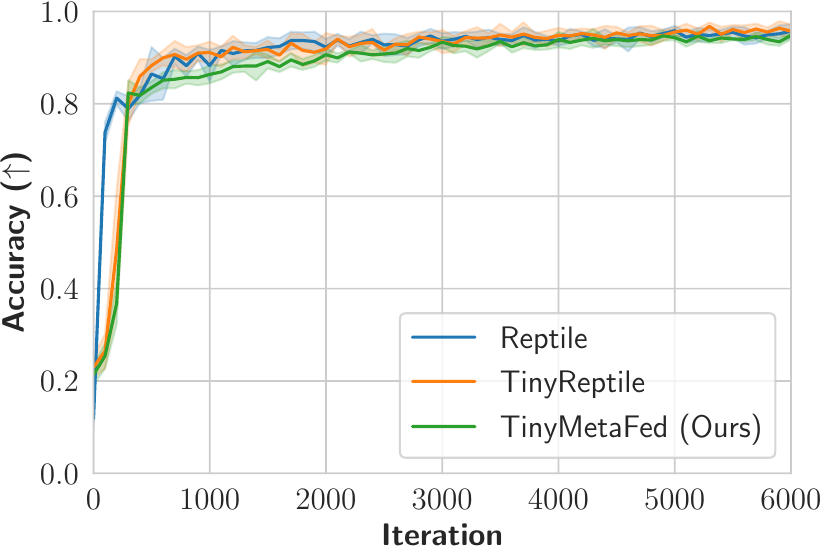}
         \caption{Omniglot 5-classes.}
         \label{fig_3_1}
     \end{subfigure}
     \hfill
     \begin{subfigure}[b]{0.48\textwidth}
         \centering
         \includegraphics[width=\textwidth]{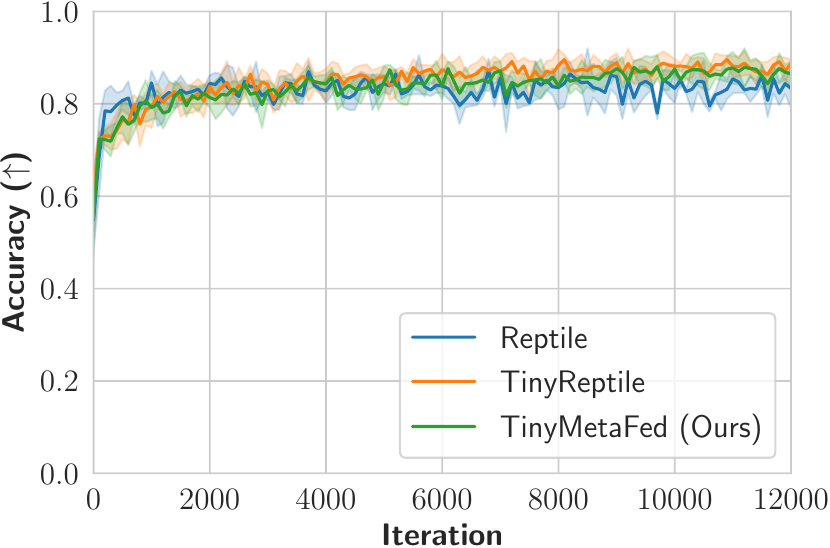}
         \caption{Keywords Spotting 3-classes.}
         \label{fig_3_2}
     \end{subfigure}
    \caption{Training convergence of Reptile, TinyReptile, and TinyMetaFed on the Omniglot and Keywords Spotting datasets. Together with the results from the Sine-wave example shown in Fig.~\ref{fig_2}, we demonstrate that TinyMetaFed can achieve similar convergence performance while significantly reducing energy consumption and communication costs.}
    \label{fig_3}
\end{figure}

We first show the convergence behaviors of the three approaches on the Sine-wave, Omniglot image classification, and Keywords Spotting audio classification datasets, as depicted in Figures~\ref{fig_2},~\ref{fig_3_1}, and~\ref{fig_3_2}, respectively. Each experiment is repeated multiple times, and the results are presented as the mean value with the standard deviation. Depending on the datasets, TinyMetaFed demonstrates similar or even better performance than the baselines. For instance, TinyMetaFed exhibits comparable final performance in the Omniglot and Keywords Spotting tasks and demonstrates faster and more stable training progress in the Sine-wave example. 

\begin{figure}[tbp]
     \centering
     \begin{subfigure}[b]{0.328\textwidth}
         \centering
         \includegraphics[width=\textwidth]{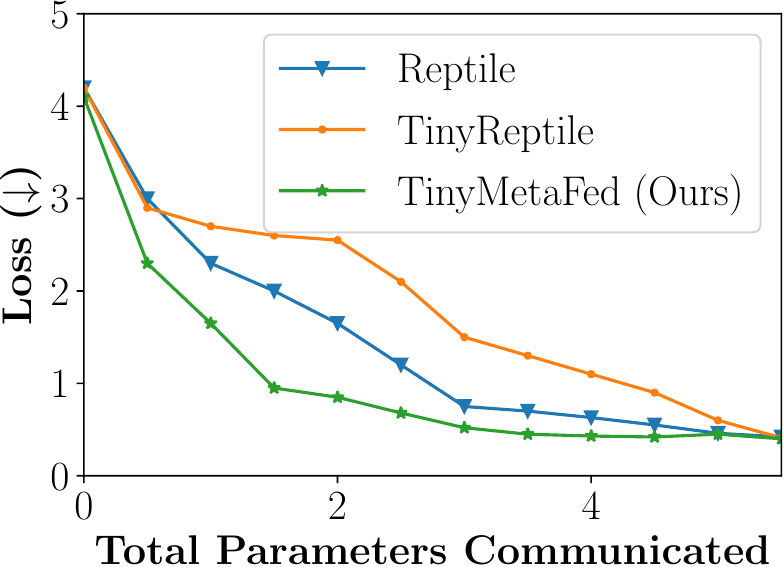}
         \caption{Sine-wave.}
     \end{subfigure}
     \hfill
     \begin{subfigure}[b]{0.328\textwidth}
         \centering
         \includegraphics[width=\textwidth]{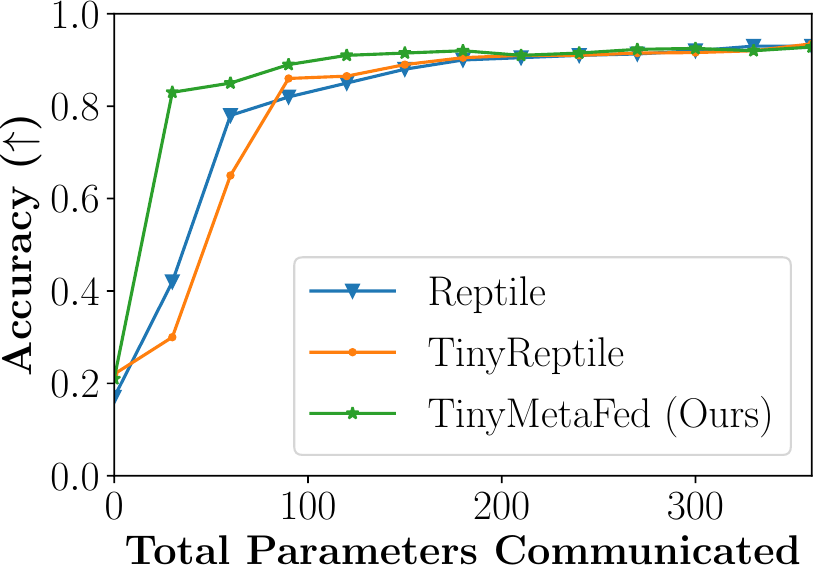}
         \caption{Omniglot 5-classes.}
     \end{subfigure}
     \hfill
     \begin{subfigure}[b]{0.328 \textwidth}
         \centering
         \includegraphics[width=\textwidth]{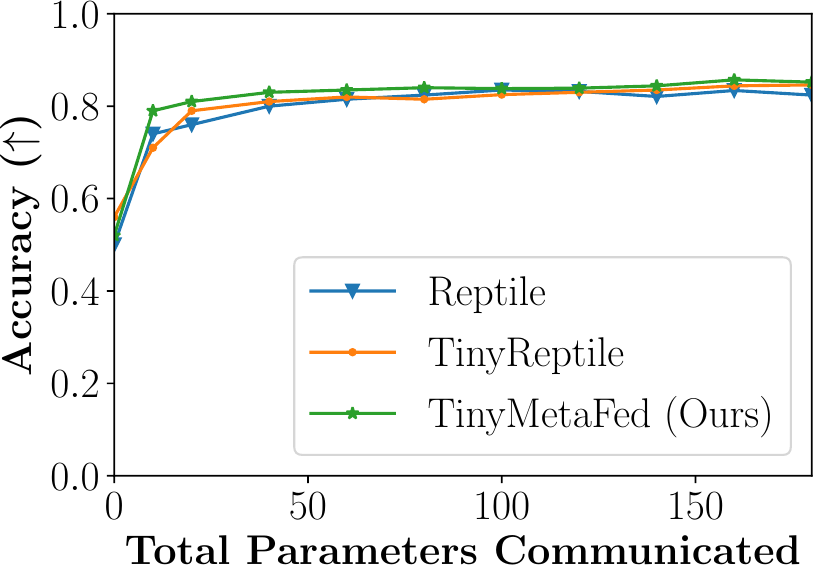}
         \caption{Keywords S. 3-classes.}
     \end{subfigure}
        \caption{Loss or accuracy as a function of total parameters communicated (in millions) between the server and one client across the datasets.}
        \label{fig_4}
\end{figure}

Next, we present the hardware benchmark results for the three tasks, as shown in Tables~\ref{tab_2},~\ref{tab_3}, and~\ref{tab_4}, respectively. We conduct the experiments on the Sine-wave example using the Arduino MCU and on the Omniglot and Keywords Spotting tasks using the Raspberry Pi. Communication costs are calculated relative to TinyReptile, which serves as the baseline with a value of one, where the entire model is transmitted to the server in each round. Since Reptile employs a batched communication schema, 
it can require communication with $N$ devices during each iteration. We measure energy consumption by subtracting idle energy consumption from the total energy consumed during algorithm execution using a USB multimeter. The results are measured on one device for one iteration. Our TinyMetaFed empirically improves upon the baseline algorithms in all metrics. For example, it achieves $65\%$ and $59\%$ communication cost savings compared to TinyReptile on the Omniglot and Keywords Spotting datasets and approximately $50\%$ energy saving on both datasets. In Fig.~\ref{fig_4}, we compare the training progress of these approaches based on the total number of parameters communicated. We observe a clear advantage of TinyMetaFed regarding communication cost saving.

Finally, TinyMetaFed provides a level of protection against privacy attacks. Many attack methods from previous work may not be effective for TinyMetaFed since only the global parameters of the model are communicated to the server, and these updates are directly calculated using a portion of the local support set. 
TinyMetaFed can be further enhanced in the most privacy-sensitive applications with secure aggregation or differential privacy to provide provable privacy guarantees.

\begin{table}[tbp]
\centering
\caption{Sine-wave: benchmark of one iteration on a Arduino Nano BLE 33.}
\label{tab_2}
\resizebox{\textwidth}{!}{%
\begin{tabular}{@{}llllllll@{}}
\toprule
 &
  Sending &
  \begin{tabular}[c]{@{}l@{}}Local Training\end{tabular} &
  Receiving &
  \begin{tabular}[c]{@{}l@{}}Total\end{tabular} &
  \begin{tabular}[c]{@{}l@{}}Communication \\ Cost\end{tabular} &
  \begin{tabular}[c]{@{}l@{}}Energy \\ Consumption\end{tabular} &
  \begin{tabular}[c]{@{}l@{}}Memory \\ Requirement\end{tabular} \\ \midrule
Reptile            & 0.81 s & 6.45 s & 0.61 s & 7.87 s & 1 * N & 2.1 J  & 8.6 KB \\
TinyReptile        & 0.81 s & 0.32 s & 0.61 s & 1.74 s & 1     & 0.48 J & 2.5 KB \\
TinyMetaFed (Ours) & \textbf{0.76 s} & \textbf{0.24 s} & \textbf{0.39 s} & \textbf{1.39 s} & \textbf{0.72}  & \textbf{0.36 J} & \textbf{2.5 KB} \\ \bottomrule
\end{tabular}%
}
\end{table}

\begin{table}[tbp]
\centering
\caption{Omniglot: benchmark of one iteration on a Raspberry Pi 4.}
\label{tab_3}
\resizebox{\textwidth}{!}{%
\begin{tabular}{@{}llllllll@{}}
\toprule
 &
  Sending &
  Local Training &
  Receiving &
  Total &
  \begin{tabular}[c]{@{}l@{}}Communication\\ Cost\end{tabular} &
  \begin{tabular}[c]{@{}l@{}}Energy \\ Consumption\end{tabular} &
  \begin{tabular}[c]{@{}l@{}}Memory \\ Requirement\end{tabular} \\ \midrule
Reptile            & 4.2 s          & 9.1 s          & 2.0 s          & 15.3 s         & 1*N           & 35 J          & 6517 KB          \\
TinyReptile        & 4.2 s          & 5.5 s          & 2.0 s          & 11.7 s         & 1             & 23 J          & 79 KB          \\
TinyMetaFed (Ours) & \textbf{1.0 s} & \textbf{5.2 s} & \textbf{0.9 s} & \textbf{7.1 s} & \textbf{0.35} & \textbf{11 J} & \textbf{79 KB} \\ \bottomrule
\end{tabular}%
}
\end{table}

\begin{table}[tbp]
\centering
\caption{Keywords Spotting: benchmark of one iteration on a Raspberry Pi 4.}
\label{tab_4}
\resizebox{\textwidth}{!}{%
\begin{tabular}{@{}llllllll@{}}
\toprule
 &
  Sending &
  Local Training &
  Receiving &
  Total &
  \begin{tabular}[c]{@{}l@{}}Communication\\ Cost\end{tabular} &
  \begin{tabular}[c]{@{}l@{}}Energy \\ Consumption\end{tabular} &
  \begin{tabular}[c]{@{}l@{}}Memory \\ Requirement\end{tabular} \\ \midrule
Reptile            & 2.7 s         & 9.6 s          & 1.5 s          & 13.8 s         & 1 * N         & 36 J         & 11462 KB        \\
TinyReptile        & 2.7 s         & 2.7 s          & 1.5 s          & 6.9 s          & 1             & 15 J         & 346 KB          \\
TinyMetaFed (Ours) & \textbf{0.7s} & \textbf{2.6 s} & \textbf{0.8 s} & \textbf{4.1 s} & \textbf{0.41} & \textbf{8 J} & \textbf{346 KB} \\ \bottomrule
\end{tabular}%
}
\end{table}

\section{Conclusion}
\label{section:conclusion}

\subsubsection{Environmental Impact}
AI has the potential to benefit society in many ways. However, the rapid development of advanced ML models in recent years has raised concerns about their sustainability and environmental impact. Efforts are underway to improve power consumption and decrease $CO_2$ emissions in ML operations. TinyML presents opportunities to enable efficient ML applications and address environmental challenges via sustainable computing practices. We believe that the future of ML is bright and tiny~\footnote{\url{https://pll.harvard.edu/course/future-ml-tiny-and-bright}}.

\subsubsection{}

This study proposes TinyMetaFed to facilitate model-agnostic meta-learning on resource-constrained tiny devices in TinyML at scale. We conduct experiments on Arduino MCU and Raspberry Pi, covering three tasks: regression, image and audio classification. Our empirical results show that TinyMetaFed can achieve significant reductions in training time (up to $60\%$), communication costs (up to $70\%$), and energy consumption (up to $50\%$) compared to the baselines. Future work includes enhancing robustness and privacy guarantee, exploring hyperparameters, and deploying in industrial use cases.

\section*{Acknowledgment}

This work is partially supported by the NEPHELE project (ID: 101070487) that has received funding from the Horizon Europe programme under the topic "Future European Platforms for the Edge: Meta Operating Systems".


%
%

%
%
%
\bibliographystyle{splncs04}
\bibliography{reference}

\begin{thebibliography}{10}
\providecommand{\url}[1]{\texttt{#1}}
\providecommand{\urlprefix}{URL }
\providecommand{\doi}[1]{https://doi.org/#1}

\bibitem{Antoniou2019}
Antoniou, A., Edwards, H., Storkey, A.: How to train your maml (2019)

\bibitem{Aramoon2021}
Aramoon, O., Chen, P.Y., Qu, G., Tian, Y.: Meta federated learning (2021)

\bibitem{Avi2022}
Avi, A., Albanese, A., Brunelli, D.: Incremental online learning algorithms
  comparison for gesture and visual smart sensors (2022)

\bibitem{Banbury2021}
Banbury, C., Reddi, V.J., Torelli, P., Holleman, J., Jeffries, N., Kiraly, C.,
  Montino, P., Kanter, D., Ahmed, S., Pau, D., Thakker, U., Torrini, A.,
  Warden, P., Cordaro, J., Di~Guglielmo, G., Duarte, J., Gibellini, S., Parekh,
  V., Tran, H., Tran, N., Wenxu, N., Xuesong, X.: Mlperf tiny benchmark (2021)

\bibitem{Dhar2021}
Dhar, S., Guo, J., Liu, J., Tripathi, S., Kurup, U., Shah, M.: A survey of
  on-device machine learning: An algorithms and learning theory perspective.
  ACM Transactions on Internet of Things  \textbf{2}(3),  1--49 (2021)

\bibitem{Finn2017}
Finn, C., Abbeel, P., Levine, S.: Model-agnostic meta-learning for fast
  adaptation of deep networks. In: Proceedings of the 34th International
  Conference on Machine Learnin. pp. 1126--1135. ICML'17, JMLR.org, Sydney,
  NSW, Australia (2017)

\bibitem{Gao2020}
Gao, Y., Kim, M., Abuadbba, S., Kim, Y., Thapa, C., Kim, K., Camtepe, S.A.,
  Kim, H., Nepal, S.: End-to-end evaluation of federated learning and split
  learning for internet of things. arXiv preprint arXiv:2003.13376  (2020)

\bibitem{Imteaj2022}
Imteaj, A., Thakker, U., Wang, S., Li, J., Amini, M.H.: A survey on federated
  learning for resource-constrained iot devices. IEEE Internet of Things
  Journal  \textbf{9}(1),  1--24 (2022)

\bibitem{Ji2021}
Ji, S., Jiang, W., Walid, A., Li, X.: Dynamic sampling and selective masking
  for communication-efficient federated learning. IEEE Intelligent Systems
  \textbf{37}(2),  27--34 (2021)

\bibitem{Jiang2023}
Jiang, B., Li, J., Wang, H., Song, H.: Privacy-preserving federated learning
  for industrial edge computing via hybrid differential privacy and adaptive
  compression. IEEE Transactions on Industrial Informatics  \textbf{19}(2),
  1136--1144 (2023)

\bibitem{Kopparapu2022}
Kopparapu, K., Lin, E., Breslin, J.G., Sudharsan, B.: Tinyfedtl: Federated
  transfer learning on ubiquitous tiny iot devices. In: 2022 IEEE International
  Conference on Pervasive Computing and Communications Workshops and other
  Affiliated Events (PerCom Workshops). pp. 79--81 (2022)

\bibitem{Lin2021}
Lin, J., Chen, W.M., Cai, H., Gan, C., Han, S.: Mcunetv2: Memory-efficient
  patch-based inference for tiny deep learning. arXiv preprint arXiv:2110.15352
   (2021)

\bibitem{LlisterriGimenez2022}
Llisterri~Giménez, N., Monfort~Grau, M., Pueyo~Centelles, R., Freitag, F.:
  On-device training of machine learning models on microcontrollers with
  federated learning. Electronics  \textbf{11}(4) (2022)

\bibitem{Loshchilov2017}
Loshchilov, I., Hutter, F.: Sgdr: Stochastic gradient descent with warm
  restarts (2017)

\bibitem{McMahan2016}
McMahan, H.B., Moore, E., Ramage, D., Hampson, S., Arcas, B.A.y.:
  Communication-efficient learning of deep networks from decentralized data.
  arXiv preprint arXiv:1602.05629  (2016)

\bibitem{Montiel2020}
Montiel, J., Halford, M., Mastelini, S.M., Bolmier, G., Sourty, R., Vaysse, R.,
  Zouitine, A., Gomes, H.M., Read, J., Abdessalem, T., Bifet, A.: River:
  Machine learning for streaming data in python (2020)

\bibitem{Nichol2018}
Nichol, A., Achiam, J., Schulman, J.: On first-order meta-learning algorithms
  (2018)

\bibitem{Parnami2022}
Parnami, A., Lee, M.: Learning from few examples: A summary of approaches to
  few-shot learning (2022)

\bibitem{Ren2021}
Ren, H., Anicic, D., Runkler, T.A.: Tinyol: Tinyml with online-learning on
  microcontrollers. In: 2021 international joint conference on neural networks
  (IJCNN). pp.~1--8. IEEE (2021)

\bibitem{Ren2023}
Ren, H., Anicic, D., Runkler, T.A.: Tinyreptile: Tinyml with federated
  meta-learning. arXiv preprint arXiv:2304.05201  (2023)

\bibitem{Rusu2018}
Rusu, A.A., Rao, D., Sygnowski, J., Vinyals, O., Pascanu, R., Osindero, S.,
  Hadsell, R.: Meta-learning with latent embedding optimization. arXiv preprint
  arXiv:1807.05960  (2018)

\bibitem{Singhal2021}
Singhal, K., Sidahmed, H., Garrett, Z., Wu, S., Rush, J., Prakash, S.:
  Federated reconstruction: Partially local federated learning. Advances in
  Neural Information Processing Systems  \textbf{34},  11220--11232 (2021)

\bibitem{Zhao2018}
Zhao, Y., Li, M., Lai, L., Suda, N., Civin, D., Chandra, V.: Federated learning
  with non-iid data. arXiv preprint arXiv:1806.00582  (2018)

\bibitem{Zhou2019}
Zhou, P., Yuan, X., Xu, H., Yan, S., Feng, J.: Efficient meta learning via
  minibatch proximal update. Advances in Neural Information Processing Systems
  \textbf{32} (2019)

\end{thebibliography}
%




\end{document}